%% file: main.tex
\documentclass{article}

\usepackage{arxiv_iclr,times}

\usepackage[utf8]{inputenc}
\usepackage[T1]{fontenc}
\usepackage{hyperref}
\usepackage{url}
\usepackage{microtype}
\usepackage{graphicx}
\usepackage{pgfplots}
\pgfplotsset{compat=1.17}
\usepackage{colortbl}
\usepackage{tabularx}
\usepackage{booktabs}
\usepackage{multirow}
\usepackage{wrapfig}
\usepackage{needspace}  
\usepackage{caption}
\usepackage{subcaption}

\usepackage{amsmath}
\usepackage{amssymb}
\usepackage{mathtools}
\usepackage{amsthm}
\usepackage{algorithm}
\usepackage{algorithmic}
\definecolor{algcmt}{HTML}{2A6F97}

\usepackage{enumitem}
\usepackage[capitalize,noabbrev]{cleveref}

\input{notation}

\title{RL Post-Training Diffusion Language Models Along Their Own Generation Paths}

\author{%
  Vaibhav Singh \\
  Mila, Concordia University
  \And
  Eugene Belilovsky \\
  Mila, Concordia University
}

\begin{document}

\maketitle
\suppressfloats[t] 

\input{sections/00_abstract}

\input{sections/01_hero_figure}

\input{sections/02_introduction}

\input{sections/03_related_work}

\input{sections/04_method}
\input{sections/05_experiments_results}

\input{sections/06_conclusion}

\subsection*{Acknowledgments}
This research was supported by NSERC. We acknowledge compute resources provided by the Digital Research Alliance of Canada).  

\bibliographystyle{iclr2027_conference}
\bibliography{papers}

\clearpage  
\appendix
\crefalias{section}{appendix}  
\input{sections/07_appendix}

\end{document}

%% file: notation.tex
\theoremstyle{plain}

\theoremstyle{definition}

\theoremstyle{remark}

\crefname{claim}{Claim}{Claims}
\Crefname{claim}{Claim}{Claims}
\crefname{conjecture}{Conjecture}{Conjectures}
\Crefname{conjecture}{Conjecture}{Conjectures}
\crefname{assumption}{Assumption}{Assumptions}
\Crefname{assumption}{Assumption}{Assumptions}

\newcommand{\Eb}{\mathbb{E}}                 

\newcommand{\KL}{\mathrm{KL}}

\definecolor{pactblue}{HTML}{2A78D6}
\definecolor{pactred}{HTML}{8A2F14}
\definecolor{gainBGa}{HTML}{DDEBFA}\definecolor{gainFGa}{HTML}{1B5AA6}   
\definecolor{gainBGb}{HTML}{DCF1ED}\definecolor{gainFGb}{HTML}{12695D}   
\definecolor{gainBGc}{HTML}{FBEEDA}\definecolor{gainFGc}{HTML}{96590A}   
\definecolor{gainBGd}{HTML}{EFE7F6}\definecolor{gainFGd}{HTML}{5D3980}   

\definecolor{gainBGbase}{HTML}{E3F1E1}\definecolor{gainFGbase}{HTML}{2C6B2C}   
\definecolor{gainBGagr}{HTML}{DDEBFA}\definecolor{gainFGagr}{HTML}{1B5AA6}      

%% file: sections/00_abstract.tex
\begin{abstract}
Diffusion language models are a promising and more efficient alternative to autoregressive models. However, most RL post-training techniques are not easily adapted to these models. One issue is that diffusion language models generate answers through many intermediate states, deciding at each step which masked positions to reveal and which tokens to place in them. Many existing RL methods used in this setting approximate intractable sequence likelihoods or weight every denoising step by the same advantage, without explicitly distinguishing how these intermediate states of the denoising trajectory affect the expected reward. To address this issue of \emph{trajectory blindness}, we introduce PACT (\textbf{P}otential-\textbf{A}nchored \textbf{C}onsistency of \textbf{T}ransitions). Motivated by a reward-tilted reference distribution, PACT jointly learns the policy and a scalar \emph{potential} that scores each intermediate state, anchored by terminal reward. For each sampled denoising transition, a consistency loss matches the policy-to-reference log-probability ratio of revealed tokens to the change in this \emph{potential}. This formulation requires neither sequence-likelihood estimation nor importance-ratio clipping. On four tasks, PACT improves LLaDA-8B-Instruct by $+\boldsymbol{9.9}$ points on GSM8K, $+\boldsymbol{12.0}$ on MATH500, $+\boldsymbol{55.7}$ on Countdown, and $+\boldsymbol{72.6}$ on Sudoku, and reaches the \textbf{best average accuracy} across the four tasks reported for this model, $\boldsymbol{+3.7}$ points over the next best method. Further compared to AGRPO, it reaches that baseline's best accuracy in about half the training compute on GSM8K.
\end{abstract}

%% file: sections/01_hero_figure.tex
%
\begin{figure}[t]
\centering
\includegraphics[width=\linewidth]{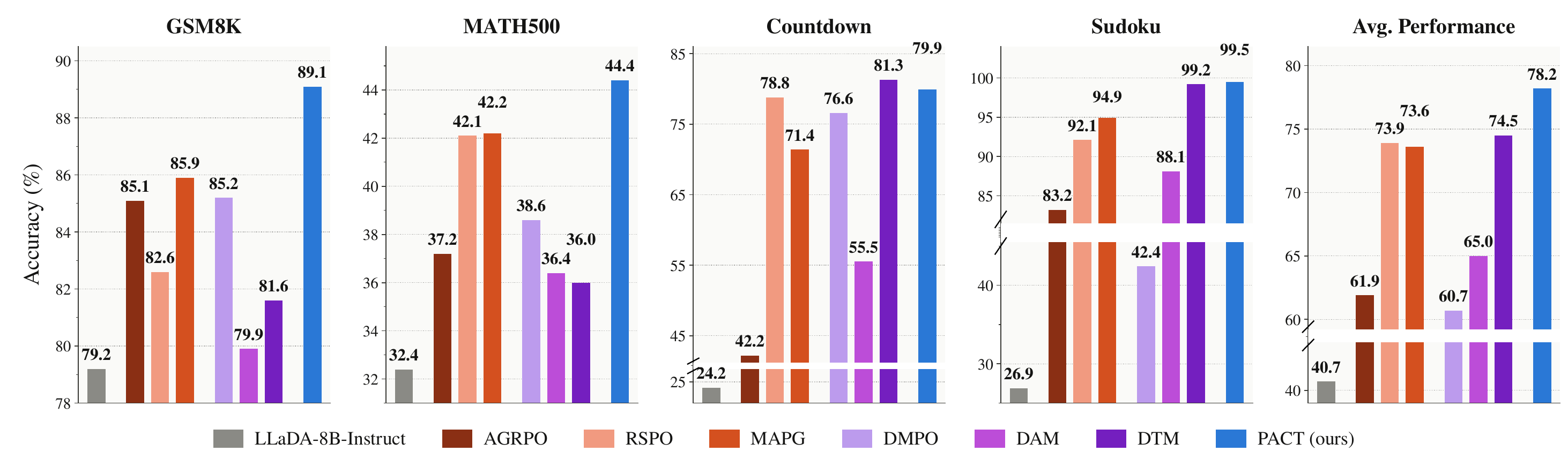}

\caption{\textbf{PACT has the best average accuracy across the four benchmarks.}
Accuracy on LLaDA-8B-Instruct at generation length 256, for three likelihood-based and three likelihood-free methods
against PACT. PACT holds the highest GSM8K, MATH500 and Sudoku accuracy, on MATH500 and Sudoku by about one
standard error; on Countdown, DTM stays ahead. The last panel is the average over the four benchmarks. Full details are mentioned in  \Cref{tab:main}.}

\label{fig:hero}
\end{figure}

%% file: sections/02_introduction.tex
\section{Introduction}\label{sec:intro}

Diffusion large language models (DLLMs) offer a promising alternative to autoregressive (AR) token generation by predicting multiple tokens in parallel \citep{nie2025large,ye2025dream,labs2025mercury, sahoo2024simple, singh2026diffumamba}. This flexibility can reduce generation latency, which is particularly useful for tasks with long reasoning horizons \citep{ye2024diffusion, svete2026reasoning, songdiffureason, singh2026fixed}. In this regard, reinforcement learning (RL) has substantially improved reasoning in AR language models \citep{shao2024deepseekmath,guo2025deepseekr1}, motivating efforts to bring similar gains to DLLMs. However, adapting these RL post-training methods to diffusion generation remains challenging.

One of the challenges is evaluating the probability of a generated response. Unlike AR models, which factorize this probability along a fixed left-to-right order, DLLMs require marginalizing over many unmasking paths. This makes computing the exact sequence likelihoods intractable. Likelihood-based RL methods address this challenge through sequence-likelihood approximations \citep{zhao2025d1,tang2025wd1,wang2025spg,yu2026relative} or formulations based on individual denoising transitions \citep{zhan2025simple,huang2025reinforcing,raajesh2026mask}. However, such methods that share a completion-level advantage apply the same weight across denoising steps, without explicitly estimating how each transition changes expected reward.

A second family of methods uses distribution matching to fit a \emph{reward-tilted} target distribution, formed by reweighting the reference distribution toward higher-reward responses. Some methods constrain completed responses \citep{ahmadi2026beyond}, while others match denoising predictions on states constructed through masking or interpolation \citep{zhu2025enhancing,chen2026tilt}. These approaches do not explicitly connect each transition along the sampler’s own path to the change in value between its intermediate states. This leaves a question: 
\emph{Can the value of an unfinished answer guide how it is completed?}

We introduce PACT (\textbf{P}otential-\textbf{A}nchored
\textbf{C}onsistency of \textbf{T}ransitions) to address this
\emph{trajectory blindness}: the absence of an explicit link
between individual denoising transitions and changes in the
potential of intermediate states. PACT jointly learns the policy and a
scalar \emph{potential} for each partial response, which scores how much reward the reference model is expected to reach when it completes that response. Terminal rewards anchor these scores. Along sampled denoising paths, a local consistency objective matches the policy-to-reference
log-probability ratio of revealed tokens to the change in
log potential between the two intermediate states of the transition. This constraint is motivated by the Doob $h$-transform \citep{nisonoff2024unlocking,bu2026dprm}, which reweights reference transitions by the ratio of their successor and current-state \emph{potentials}. PACT requires neither sequence-likelihood estimation nor importance-ratio clipping. We evaluate PACT on four tasks: GSM8K \citep{cobbe2021training}, MATH500 \citep{hendrycks2021measuring,lightman2023let}, Countdown \citep{pan2025tinyzero}, and Sudoku \citep{wang2025spgrepo}. PACT improves over the base LLaDA-8B-Instruct on all four benchmarks and has the best average accuracy among DLLM post-training methods. 

Our contributions are summarized below.
\begin{itemize}[leftmargin=1.35em,itemsep=1pt,topsep=2pt]

 \item 
    We identify that the reward of a completed answer can hide inconsistencies between the denoising steps that produced it, as transitions that raise and transitions that lower the expected reward of the intermediate states can cancel. Training each transition on its own change in expected reward, rather than the answer as a whole, is what sustains learning (\cref{sec:ablations}).
     
  \item 
  To this end, we introduce PACT, which learns reward-anchored \emph{potential} for each intermediate state and uses changes in this potential to train individual denoising transitions. This provides local training targets without sequence-likelihood estimation. 
   \item \textbf{SOTA on average across four reasoning tasks.} On LLaDA-8B-Instruct, PACT achieves the best average accuracy among DLLM post-training methods across the four benchmarks, \textbf{$+3.7$ points over the next best method} at length 256, and improves over the base model by $+9.9$ to $+72.6$ points. It also matches AGRPO's best GSM8K accuracy in about half the training compute.
\end{itemize}

%% file: sections/03_related_work.tex
\section{Related Work}\label{sec:related}

\textbf{Likelihood-based reward maximization.} 
Sequence-likelihood surrogates make policy optimization tractable for DLLMs, as in d1, wd1, SPG, and RSPO \citep{zhao2025d1,tang2025wd1,wang2025spg,yu2026relative}.
Trajectory-based methods instead use denoising action probabilities: AGRPO samples transitions to estimate its objective, while LLaDOU and MAPG also optimize position selection \citep{zhan2025simple,huang2025reinforcing,raajesh2026mask}. These methods already learn from intermediate decisions. PACT adds an explicit constraint linking each sampled transition to a change in learned potential for intermediate states. wd1++, an extension in the revised wd1 paper, also scores the model's predicted completions at intermediate denoising steps with the group advantage \citep{tang2025wd1}, but does not learn how the potential changes between successive states.

\textbf{Likelihood-free distribution matching.} Distribution matching offers another route to reward-guided post-training. DMPO uses reward-weighted denoising losses, while DTM matches tilted unmasking posteriors on constructed intermediate states \citep{zhu2025enhancing,chen2026tilt}. Other likelihood-free approaches match transition rates to
discrete-adjoint estimates or distill a reward-guided denoiser \citep{so2026discrete,tang2026gdsd}.
These methods already incorporate reward information into local predictions. PACT instead learns a scalar potential alongside the policy and constrains its change across recorded denoising transitions. This provides a direct relation between the policy's preference for a sampled transition and its predicted improvement in the partial response.

\textbf{Flow balance, relative to a reference.}
PACT's per-transition condition follows from the Doob $h$-transform of the reference sampler (\cref{sec:method}). The same condition can also be interpreted in the flow-balance framework. GFlowNets learn to sample from a target distribution through
balance constraints, imposed either across individual
transitions (\emph{detailed balance}) or over complete trajectories (\emph{trajectory balance})  \citep{bengio2021flow,malkin2022trajectory}. TraFL, FlowRL and GFlowRL use trajectory-balance objectives for language-model post-training with reward-tilted targets \citep{ahmadi2026beyond,zhu2025flowrl,liu2026gflowrl}. Related methods learn from intermediate states:
$\nabla$-GFN uses reward gradients to fine-tune diffusion models through local balance constraints relative to a pretrained reference \citep{liu2024diversity}, while DDPP matches denoising posteriors, with a variant that learns a state-dependent normalizer \citep{rectorbrooks2024steering}. In this flow-balance view, PACT builds on these ideas through a reference-relative
detailed-balance condition: each sampled transition's
policy-to-reference probability ratio must equal the ratio of its successor and current-state potentials, which PACT learns from terminal rewards and from local consistency along the sampled denoising trajectory. At the residual's fixed point the potential of the fully masked state equals $\log Z_q$, so no separate normalizer network is fitted. The same condition appears in KL-regularized RL as one-step path consistency \citep{nachum2017bridging,nachum2018trust}, with the value $V=\beta\log h$ in place of the potential, and relative trajectory balance reduces to it as well \citep{deleu2025relative}. What PACT adds is a reward-anchored potential in place of a fitted value or normalizer, applied to a DLLM's own sampled transitions.

%% file: sections/04_method.tex
\section{Method: Potential-Anchored Consistency}\label{sec:method}

\textbf{Denoising trajectories.} Given a prompt $q$, a DLLM generates a response along a trajectory $\tau=(x_T,\ldots,x_0)$, from a masked answer window $x_T$ to the completed response $x_0=y$. A step commits a subset $S_t$ of the still-masked slots and writes a token into each; conditioned on that choice, the committed tokens are independent, so the transition is
\begin{equation}\label{eq:kernel}
P_\theta(x_{t-1}\mid x_t, S_t) \;=\; \prod_{i\in S_t}\pi_\theta(a_i \mid x_t, i),
\end{equation}
where $\pi_\theta(\cdot \mid x_t, i)$ is the model's token distribution at slot $i$, read from the step's single forward pass, and $a_i$ the token committed there. The sequence marginal $p_\theta(y \mid q) = \sum_{\tau \to y} p_\theta(\tau \mid q)$, by contrast, sums over every commitment order and is intractable. PACT utilizes local token probabilities as given in \cref{eq:kernel} and not the marginal. Which slots form $S_t$ is drawn from the policy's own confidences $c_\theta(x_t,i)=\max_a\log\pi_\theta(a\mid x_t,i)$, one slot at a time without replacement, at a temperature $\tau_o$; $\tau_o\to0$ recovers confidence decoding, and $\tau_o>0$ lets the sampler explore orders it would otherwise never try. A full transition is the order draw followed by the token draw of \cref{eq:kernel}; the policy learns the order only implicitly, through its confidences, and the loss does not train it (\cref{sec:ablations}).

\begin{figure}[t]
\centering
\includegraphics[width=\linewidth]{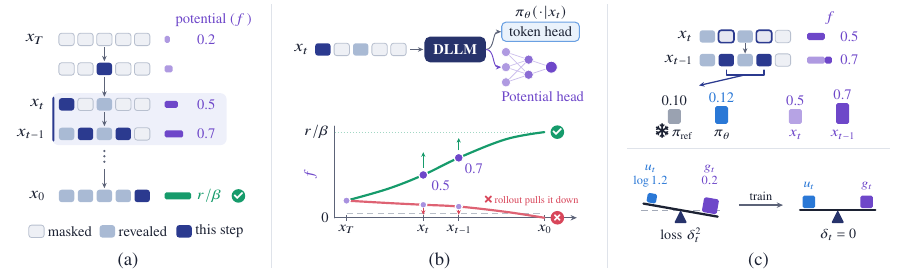}
\caption{\small \textbf{Overview of PACT training.}
\textbf{(a)}~PACT scores a few transitions of each sampled path; every state has a \textbf{potential} $f=\log h$, equal to $r/\beta$ at the answer and $0$ after a failed one.
\textbf{(b)}~A small \textbf{potential head} predicts $f$ from the DLLM's hidden state, pulled up by correct rollouts and down by failed ones.
\textbf{(c)}~On each scored step, PACT \textbf{minimizes {\boldmath$\delta_t^2=(u_t-g_t)^2$}}, where $u_t$ is the revealed tokens' log-ratio to the frozen reference and $g_t$ the gain in log potential (\cref{eq:loss}).
Numbers are illustrative.}
\label{fig:overview}
\end{figure}

\textbf{Post-training as a reward-tilted target.} Let $r(y) \in [0,1]$ be a verifiable terminal reward on the finished text, and let $p_{\mathrm{ref}}$ be a frozen reference, the base model with adapters disabled, whose transitions draw the order the same way from its own confidences and the tokens from \cref{eq:kernel}. Over unrestricted trajectory distributions supported by the reference, the objective $\max_p\,\Eb_p[r]-\beta\,\KL(p\|p_{\mathrm{ref}})$ has optimum \citep{levine2018reinforcement}
\begin{equation}\label{eq:tilt}
p^\star(\tau \mid q) \;=\; \frac{1}{Z_q}\, p_{\mathrm{ref}}(\tau \mid q)\, e^{\,r(y)/\beta},
\qquad
Z_q = \textstyle\sum_\tau p_{\mathrm{ref}}(\tau \mid q)\, e^{\,r(y)/\beta}.
\end{equation}
Smaller $\beta$ favors high-reward responses more strongly. Marginalizing over trajectories recovers the reward-tilted response distribution discussed in \cref{sec:intro}, while preserving the reference distribution over paths conditional on the completed response.

\textbf{PACT objective: From terminal rewards to local consistency.} Let the \emph{potential} of a state be defined as,
\begin{equation}\label{eq:potential}
h(x) \;=\; \Eb_{p_{\mathrm{ref}}}\!\big[\, e^{\,r(y)/\beta} \,\big|\, x \,\big].
\end{equation}
It measures the expected exponential reward of completing the partial response with the reference model. At the endpoints, $h(x_0)=e^{r(y)/\beta}$ and $h(x_T)=Z_q$. Since denoising is a Markov process, the reward tilt reweights each reference transition by the successor-to-current potential ratio, giving the Doob $h$-transform \citep{nisonoff2024unlocking,bu2026dprm}: 
\begin{equation}\label{eq:identity}
P^\star(x' \mid x) \;=\; P_{\mathrm{ref}}(x' \mid x)\,\frac{h(x')}{h(x)},
\quad\text{i.e.}\quad
\log\frac{P^\star(x' \mid x)}{P_{\mathrm{ref}}(x' \mid x)} \;=\; \log\frac{h(x')}{h(x)}.
\end{equation}
Thus, the ideal policy changes each reference transition's probability according to the potential gained by that transition. This identity motivates PACT's local consistency objective.

PACT enforces \cref{eq:identity} on the transitions the sampler took, learning the policy $\theta$ and a potential head $f_\phi(x)=\log h_\phi(x)$ that reads the policy's pooled hidden state; the reward enters only through the head, as the boundary anchor $f_\phi(x_0)\coloneqq r(y)/\beta$ and as the target of the calibration term below. Over trajectories $\tau\sim\pi_\theta(\cdot\mid q)$ it minimizes the loss $\mathcal{L}(\theta,\phi)$
\begin{equation}\label{eq:loss}
\mathcal{L}=\Eb_{\tau\sim\pi_\theta}\Eb_{t\sim\tau}\Bigl[
\underbrace{\left(\sum_{i\in S_t}\log\!\frac{\pi_\theta(a_i\mid x_t,i)}{\pi_{\mathrm{ref}}(a_i\mid x_t,i)}-f_\phi(x_{t-1})+f_\phi(x_t)\right)^2}_{\text{consistency residual }\delta_t^2=(u_t-g_t)^2}
+\underbrace{\sum_{x\in\mathcal{X}_t}\bigl(f_\phi(x)-\tfrac{r(y)}{\beta}\bigr)^2}_{\text{reward anchors}}
\Bigr].
\end{equation}
The consistency term $(u_t-g_t)^2$ links the revealed-token log-ratio to the change in log potential. The reward anchors, over the transition's two states $\mathcal{X}_t=\{x_t,x_{t-1}\}$, tie intermediate scores to the observed terminal reward and vanish at the terminal state by construction. We use the scaled reward $r(y)/\beta$ as a practical regression target because estimating the ideal potential directly would require additional reference-model completions from each state or importance weighting, while exponential reward targets can amplify variance at small $\beta$. This choice makes the anchor a practical value target rather than an unbiased estimator of the ideal log potential. Specifically, under the rollout distribution $p_{\mathrm{roll}}$, Jensen's inequality gives $\Eb_{p_{\mathrm{roll}}}[r(Y)/\beta\mid x]\leq\log\Eb_{p_{\mathrm{roll}}}[e^{r(Y)/\beta}\mid x]$; the anchor alone targets the left-hand side, whereas the ideal log potential would have used the reference distribution. At the anchors' optimum $f_\phi$ is a Monte-Carlo estimate of the rollout policy's expected scaled reward, a value function, and the residual asks each step's token log-ratio to match the increment of that value: a one-step path-consistency condition of KL-regularized RL \citep{nachum2017bridging,nachum2018trust}; with the order term scored, the residual alone has the $h$-transform of \cref{eq:identity} as its fixed point, and the anchors move the minimizer toward the value fit. Combined with local consistency, this objective requires neither sequence-likelihood estimation nor policy importance-ratio clipping, and still yields a consistency gradient, if a small one, on prompts whose completions all receive the same reward, where a group-relative advantage is exactly zero.

\textbf{Implementation.} We learn the potential with a potential head, a two-layer MLP that maps the mean-pooled final hidden state of the answer window to $f_\phi(x)$ (\cref{fig:overview}(b)). The head reads the representation the policy already computes, so $f_\phi(x_t)$ comes from the same forward pass as $\pi_\theta(\cdot \mid x_t)$, and the successor score $f_\phi(x_{t-1})$ needs one more. The expectation over transitions is estimated with $K$ transitions per trajectory, stratified across its steps (\cref{alg:pact}). Gradients flow through the head into the policy adapters. The reward anchors regress both scores toward the rollout's scaled reward $r(y)/\beta$ (\cref{alg:pact}, line~\ref{alg:line-anchor}), while the consistency residual brings $u_t$ and $g_t$ into agreement. At zero residual, the revealed-token probability ratio equals $h_\phi(x_{t-1})/h_\phi(x_t)$. In \cref{fig:overview}(c) the gain in log potential exceeds the token log-ratio, so $\delta_t<0$: the residual pushes $u_t$ up and $g_t$ down, with the actual update also shaped by the anchors and the shared parameters. The head is used only in training; decoding uses the policy alone. Rollout and optimizer settings appear in \cref{app:configs}.

\begin{algorithm}[t]
\caption{PACT training}\label{alg:pact}
\begin{algorithmic}[1]
\REQUIRE policy $\pi_\theta$, potential head $f_\phi=\log h_\phi$, reward $r$, prompts $\mathcal{D}$; tilt $\beta$, rollouts per prompt $N$, transitions scored per rollout $K$, training steps $M$
\ENSURE trained policy $\pi_\theta$ and potential head $f_\phi$
\STATE $\pi_{\mathrm{ref}} \leftarrow \pi_\theta$ with adapters disabled
\FOR{step $= 1$ to $M$}
  \STATE sample prompt batch $\mathcal{D}_b \sim \mathcal{D}$; \; $\mathcal{L} \leftarrow 0$
  \FORALL{$q \in \mathcal{D}_b$ and $n = 1,\dots,N$}
    \STATE sample $\tau = (x_T \to \cdots \to x_0)$ from $\pi_\theta$ on $q$; \; $f_\phi(x_0) \coloneqq r(y)/\beta$ \COMMENT{boundary anchor}
    \FOR{$t \in \mathcal{T}_K$, $K$ steps stratified up to the last answer token}
      \STATE $u_t \leftarrow \sum_{i \in S_t}\big[\log \pi_\theta - \log \pi_{\mathrm{ref}}\big](a_i \mid x_t, i)$ \COMMENT{log-ratio of the step}
      \STATE $g_t \leftarrow f_\phi(x_{t-1}) - f_\phi(x_t)$ \COMMENT{potential gained by the step}\label{alg:line-gain}
      \STATE $\delta_t \leftarrow u_t - g_t$ \COMMENT{consistency residual}
      \STATE $\mathcal{L}_{sum} \leftarrow \mathcal{L}_{sum} + \delta_t^{\,2} + \sum_{x \in \{x_t,\,x_{t-1}\}}\big(f_\phi(x) - r(y)/\beta\big)^2$ \COMMENT{reward anchor}\label{alg:line-anchor}
    \ENDFOR
  \ENDFOR
  \STATE $\hat{\mathcal{L}}\leftarrow \frac{\mathcal{L}_{sum}}{|\mathcal{D}_b|NK}$
  
  \STATE \text{Take an optimizer step minimizing} $\hat{\mathcal{L}}$ with respect to $(\theta, \phi)$ \COMMENT{\cref{eq:loss}}
\ENDFOR
\end{algorithmic}
\end{algorithm}

\textbf{Relation to trajectory balance.} Summing the residuals of every step of a trajectory, scored with the full kernel, order and tokens, cancels the intermediate potentials:
\begin{equation}\label{eq:telescope}
\sum_t \delta_t \;=\; \log\frac{p_\theta(\tau\mid q)}{p_{\mathrm{ref}}(\tau\mid q)} \;-\; \frac{r(y)}{\beta} \;+\; \log h_\phi(x_T).
\end{equation}
This is the relative trajectory-balance residual \citep{venkatraman2024amortizing} with $\log h_\phi(x_T)$ in place of $\log Z_q$, and at the fixed point of the residual $h_\phi(x_T) = Z_q$. A whole-path residual can be small even when local inconsistencies are large, because errors across transitions can cancel. PACT penalizes each sampled residual separately, directly constraining how a transition's log-ratio relates to its change in potential. TraFL \citep{ahmadi2026beyond} instead uses masked-reconstruction likelihood surrogates for completed responses and learns a prompt-dependent normalizer. PACT avoids this sequence-likelihood approximation by scoring revealed tokens on recorded transitions, while the same potential head supplies both intermediate scores and the initial normalizer. PACT outperforms TraFL in our comparison under shared rollout settings (\cref{app:trafl}).

%% file: sections/05_experiments_results.tex
\section{Experiments}\label{sec:experiments}

\subsection{Training details}\label{sec:training}
We apply PACT on LLaDA-8B-Instruct \citep{nie2025large} with no supervised fine-tuning stage. We use LoRA adapters 
with $r=128$ where each run sees 900 rollout batches of 128 completions, 16 prompts with 8 rollouts apiece according to AGRPO's released schedule \citep{zhan2025simple}. PACT trains at the geometry it is evaluated at, generation length 256 in semi-autoregressive blocks of 32 over 128 denoising steps, rolling out under stochastic-confidence order at temperature $\tau_o = 0.2$ with the reference tilt set to $\beta = 0.1$. The potential head $h_\phi$ is a two-layer MLP of hidden width 256 read from the mean-pooled final hidden state of the answer window, zero-initialized and given its own learning rate and gradient clip so its gradients never draw on the policy's clip budget. The policy learning rate is $2\times10^{-5}$ on GSM8K and $5\times10^{-6}$ on the other tasks, and complete hyperparameters are in \cref{app:configs}.

\begin{figure}[t]
\centering
\includegraphics[width=\linewidth]{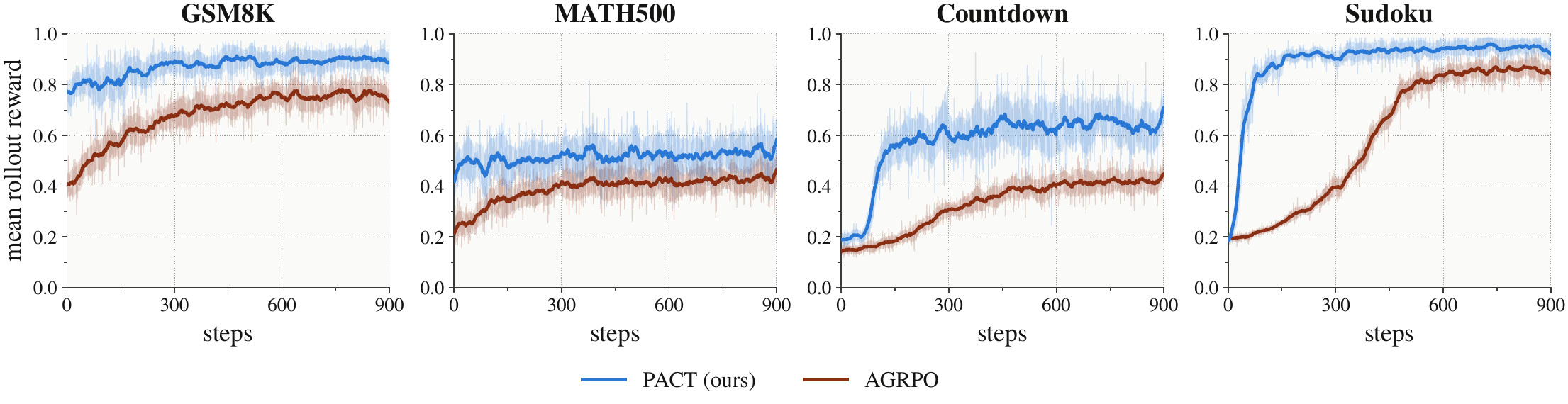}
\caption{\small \textbf{PACT's rollout reward leads AGRPO's on all four tasks throughout training.} Mean rollout reward against training step,
for all four tasks against AGRPO trained on the same rollout budget, rescaled to $[0,1]$. The faint trace is the per-step value as logged, the solid line a
25-step moving average, the band $\pm1$ rolling standard deviation over the same window. On every task PACT's moving average lies above AGRPO's for the whole run; the gap is widest on Sudoku, where PACT
reaches $0.90$ by step 291 while AGRPO stands at $0.39$ at that step and reaches $0.84$ only at step 900.}
\label{fig:rewards}
\vspace{-5pt}
\end{figure}

\subsection{Datasets}\label{sec:datasets}
We train PACT on four tasks. GSM8K \citep{cobbe2021training} poses grade-school arithmetic word problems that take a few steps of reasoning; we train on its standard split, and the reward is a format term plus exact match on the final answer. MATH500 is competition mathematics: we train on the 7{,}500-problem MATH training split \citep{hendrycks2021measuring}, evaluate on the 500-problem MATH-500 subset \citep{lightman2023let}, and reuse the GSM8K reward. Countdown \citep{pan2025tinyzero} gives three integers and a target and asks for an arithmetic expression that reaches it, and the reward is a format term plus a verifier that checks both the arithmetic and the use of each number. Sudoku \citep{wang2025spgrepo} asks the model to complete a $4{\times}4$ puzzle, where the reward is a format term plus the fraction of initially empty cells filled correctly. We rescale every reward to $[0,1]$ and follow the split lineage of d1 \citep{zhao2025d1} and SPG \citep{wang2025spg} where it exists.

\subsection{Evaluation}\label{sec:evalproto}

\textbf{Protocol.} Every number we measure is greedy Pass@1: one sample per prompt, decoded at temperature $0$ in low-confidence order, in blocks of 32 with the denoising steps set to half the generation length, matching AGRPO's evaluation \citep{zhan2025simple}. We apply this one protocol to every checkpoint we evaluate, ours and the reproduced baselines, at generation lengths 256 and 512 (\cref{tab:main}). We evaluate a checkpoint every 25 rollout batches and report the best test accuracy among them, for PACT and the reproduced baselines alike, as AGRPO (best of its last 150 steps), d1, SPG and RSPO do \citep{zhan2025simple,zhao2025d1,wang2025spg,yu2026relative}.

\textbf{Baselines.} We compare PACT against the published DLLM post-training methods of \cref{tab:main}, in the two families \cref{sec:related} describes: the likelihood-based methods d1 \citep{zhao2025d1}, wd1 \citep{tang2025wd1}, UniGRPO \citep{yang2025mmada}, SPG \citep{wang2025spg}, RSPO \citep{yu2026relative}, MAPG \citep{raajesh2026mask}, and AGRPO \citep{zhan2025simple}; and the likelihood-free methods DMPO \citep{zhu2025enhancing}, DAM \citep{so2026discrete}, and DTM \citep{chen2026tilt}. A fair comparison needs a matched budget, so for the three methods we could reproduce from public code (AGRPO, wd1, and DMPO) we retrain each ourselves for the same 900 rollout rounds and decode it under this protocol; those rows carry a $^{\dagger}$ in \cref{tab:main}, and every other row is each paper's own self-report.

\section{Results}\label{sec:results}

Under the same rollout budget as AGRPO, PACT improves on the base model on all four benchmarks (\cref{tab:main}): $+9.9$ on GSM8K (89.1 against 79.2), $+12.0$ on MATH500 (44.4 against 32.4), $+55.7$ on Countdown (79.9 against 24.2), and $+72.6$ on Sudoku (99.5 against 26.9), the largest gains falling on Countdown and Sudoku, where the base model is weakest ($24.2$ and $26.9$).
\Needspace*{16\baselineskip}
\begin{wrapfigure}{r}{0.49\linewidth}
\vspace{-12pt}
\centering
\includegraphics[width=\linewidth]{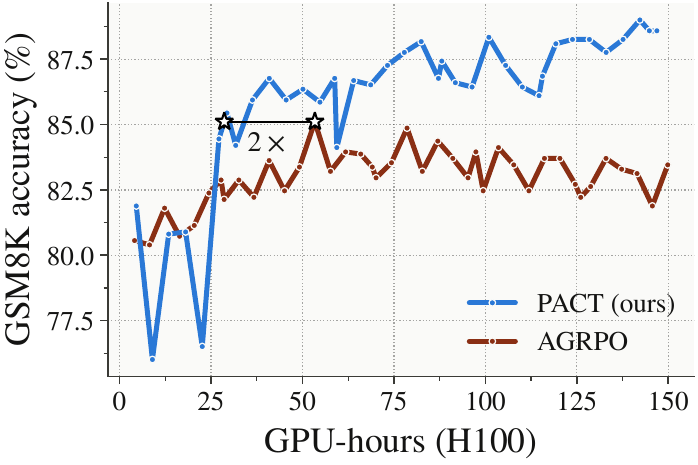}
\caption{\small \textbf{PACT reaches AGRPO's best GSM8K accuracy in about half the compute.} Held-out accuracy at generation length 256 against training GPU-hours (16 H100s), at every evaluated checkpoint up to a shared budget of $150$ GPU-hours. The stars mark AGRPO's best checkpoint, $85.1$ after $53.4$ GPU-hours, and the point where PACT first reaches it, after $28.6$: about $2\times$ saving. At matched compute, PACT leads by $2.6$, $4.0$ and $5.1$ points at $50$, $100$ and $150$ GPU-hours.}
\label{fig:wallclock}
\vspace{-5pt}
\end{wrapfigure}
Against the published methods, PACT has the best average accuracy at both lengths, 78.2 against DTM's 74.5 at length 256. On GSM8K its 89.1 clears the strongest baseline, SPG's 86.1, by three points, about three standard errors of the test accuracy; on MATH500, 44.4 against MAPG's 42.2 is about one standard error (2.2 points), and on Sudoku 99.5 against DTM's 99.2 is within one puzzle; on Countdown DTM's 81.3 leads our 79.9 at length 256, and our 80.7 is the highest at 512.

\begin{table}[t]
\centering
\caption{\small \textbf{PACT has the best average accuracy at both lengths.}
Accuracy (\%) on LLaDA-8B-Instruct at lengths 256 and 512; \textbf{bold marks the best value in each column}, underline the second best, and the coloured offset is PACT's margin over the best other method (blue ahead, red behind).
PACT holds the largest value in every column but Countdown at 256, where DTM leads, and ties DTM on Sudoku at 512; the MATH500 and Sudoku margins are within one standard error.
$^{\dagger}$\,Our reproduction with the authors' code, 900 rollout rounds, decoded as in \cref{sec:evalproto}. Sudoku is 3-shot on SPG's split in every row but RSPO and DAM (zero-shot).}
\label{tab:main}
\small
\setlength{\tabcolsep}{3pt}
\begin{tabularx}{\linewidth}{l *{10}{>{\centering\arraybackslash}X}}
\toprule
 & \multicolumn{2}{c}{GSM8K} & \multicolumn{2}{c}{MATH500} & \multicolumn{2}{c}{Countdown}
 & \multicolumn{2}{c}{Sudoku} & \multicolumn{2}{c}{\textbf{Average}} \\
\cmidrule(lr){2-3}\cmidrule(lr){4-5}\cmidrule(lr){6-7}\cmidrule(lr){8-9}\cmidrule(lr){10-11}
 & 256 & 512 & 256 & 512 & 256 & 512 & 256 & 512 & 256 & 512 \\
\midrule
Base & 79.2 & 79.8 & 32.4 & 34.6 & 24.2 & 30.9 & 26.9 & 18.5 & 40.7 & 41.0 \\
\midrule
\multicolumn{11}{l}{\emph{Likelihood-based methods}}\\
d1 \citep{zhao2025d1} & 81.1 & 82.1 & 38.6 & 40.2 & 32.0 & 42.2 & 32.5 & 29.3 & 46.1 & 48.5 \\
wd1$^{\dagger}$ \citep{tang2025wd1} & 84.4 & 85.4 & 38.6 & 41.4 & 60.9 & 58.6 & 32.1 & 22.5 & 54.0 & 52.0 \\
SPG \citep{wang2025spg} & \underline{86.1} & 84.5 & 40.0 & 41.8 & 70.7 & 70.3 & 94.0 & 93.1 & 72.7 & 72.4 \\
UniGRPO \citep{yang2025mmada} & 82.5 & 82.7 & 37.4 & 39.4 & 43.0 & 57.0 & 67.0 & 62.9 & 57.5 & 60.5 \\
RSPO \citep{yu2026relative} & 82.6 & 80.6 & 42.1 & 40.3 & 78.8 & 73.8 & 92.1 & 90.8 & 73.9 & 71.4 \\
MAPG \citep{raajesh2026mask} & 85.9 & \underline{87.1} & \underline{42.2} & \underline{44.2} & 71.4 & 71.0 & 94.9 & \underline{96.2} & 73.6 & 74.6 \\
AGRPO$^{\dagger}$ \citep{zhan2025simple} & 85.1 & 86.3 & 37.2 & 41.4 & 42.2 & 51.6 & 83.2 & 86.8 & 61.9 & 66.5 \\
\midrule
\multicolumn{11}{l}{\emph{Likelihood-free methods}}\\
DMPO$^{\dagger}$ \citep{zhu2025enhancing} & 85.2 & 83.6 & 38.6 & 42.6 & 76.6 & \underline{80.1} & 42.4 & 30.6 & 60.7 & 59.2 \\
DAM \citep{so2026discrete} & 79.9 & 77.0 & 36.4 & 37.2 & 55.5 & 59.0 & 88.1 & 84.9 & 65.0 & 64.5 \\
DTM \citep{chen2026tilt} & 81.6 & 83.2 & 36.0 & 40.2 & \textbf{81.3} & 78.9 & \underline{99.2} & \textbf{99.4} & \underline{74.5} & \underline{75.4} \\
\midrule
\rowcolor{black!8}
PACT (ours) & \begin{tabular}[c]{@{}c@{}}\textbf{89.1}\\[-2pt]{\color{pactblue}\scriptsize+3.0}\end{tabular} & \begin{tabular}[c]{@{}c@{}}\textbf{89.0}\\[-2pt]{\color{pactblue}\scriptsize+1.9}\end{tabular} & \begin{tabular}[c]{@{}c@{}}\textbf{44.4}\\[-2pt]{\color{pactblue}\scriptsize+2.2}\end{tabular} & \begin{tabular}[c]{@{}c@{}}\textbf{45.8}\\[-2pt]{\color{pactblue}\scriptsize+1.6}\end{tabular} & \begin{tabular}[c]{@{}c@{}}\underline{79.9}\\[-2pt]{\color{pactred}\scriptsize$-$1.4}\end{tabular} & \begin{tabular}[c]{@{}c@{}}\textbf{80.7}\\[-2pt]{\color{pactblue}\scriptsize+0.6}\end{tabular} & \begin{tabular}[c]{@{}c@{}}\textbf{99.5}\\[-2pt]{\color{pactblue}\scriptsize+0.3}\end{tabular} & \begin{tabular}[c]{@{}c@{}}\textbf{99.4}\\[-2pt]{\color{pactblue}\scriptsize+0.0}\end{tabular} & \begin{tabular}[c]{@{}c@{}}\textbf{78.2}\\[-2pt]{\color{pactblue}\scriptsize+3.7}\end{tabular} & \begin{tabular}[c]{@{}c@{}}\textbf{78.7}\\[-2pt]{\color{pactblue}\scriptsize+3.3}\end{tabular} \\
\bottomrule
\end{tabularx}
\end{table}

We observe that PACT's improvement cannot be attributed to extra compute. In all our experiments above where we reproduced the baselines we compare them at equal rollouts, but each batch in PACT costs more: per sampled transition PACT adds two gradient forward passes and one reference pass without gradient, so a batch takes $41.3$ s against AGRPO's $37.5$ on 16 H100s, and the 900 batches $165$ against $150$ GPU-hours. Read at equal compute, \cref{fig:wallclock} puts PACT at AGRPO's best GSM8K accuracy after $28.6$ GPU-hours, where AGRPO itself needed $53.4$, about $2\times$ less; the crossing depends on checkpoint noise, so we also compare at matched compute, where PACT leads by $2.6$, $4.0$ and $5.1$ points at $50$, $100$ and $150$ GPU-hours.

We see two reasons for the gain. First, every quantity in the residual is computed exactly. The kernel factorizes into an order draw and the token softmaxes at the positions that draw commits (\cref{eq:kernel}), and \cref{eq:loss} evaluates it only at the transition that actually occurred, so no marginal over unmasking orders is ever needed. The likelihood-based methods that score a completed answer do need that marginal and must approximate it, and where a ratio survives it is clipped, so the approximation's error enters the gradient.

Second, the reward enters only through the potential head as the boundary value $\log h_\phi(x_0) = r(y)/\beta$ and the calibration anchor that holds $h_\phi$ near it, so the policy never sees $r$ as a per-token signal. The residual pulls each transition's log-ratio to the reference toward the change in potential across that step, so a correct answer does not raise the probability of its whole trajectory evenly: it raises the steps that increase the potential and keeps steps with no gain close to the reference, where a policy gradient multiplies every token by one scalar advantage. Because the residual scores the single order the sampler took rather than a marginal over orders, that credit is assigned in the order the tokens were committed. The order is drawn from the policy's own confidences, but the default token-only residual does not train it. \Cref{sec:ablations} studies both.

\begin{figure}[t]
\centering
\includegraphics[width=\linewidth]{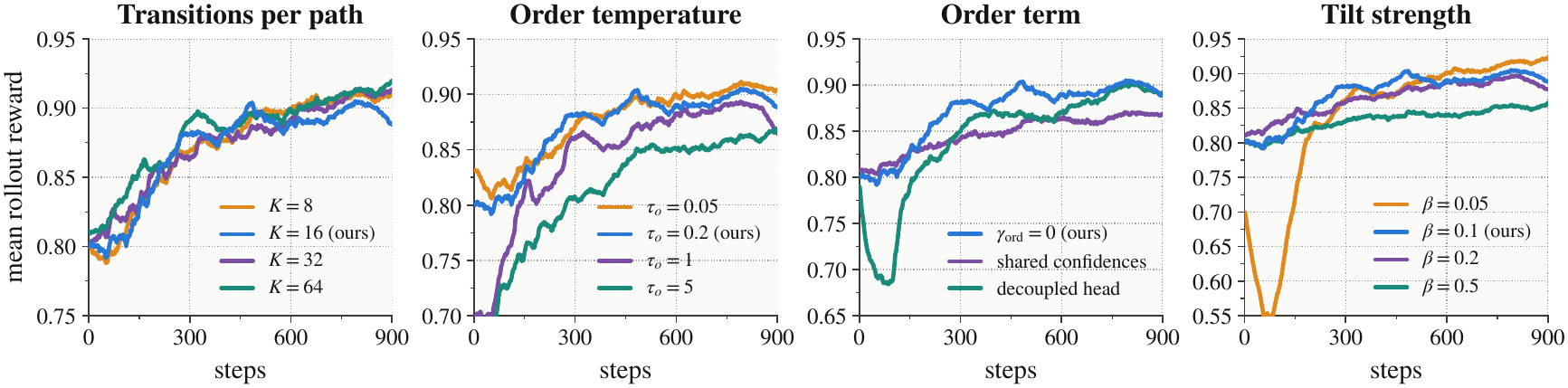}
\caption{\small \textbf{Ablations: training reward.} Mean rollout reward on GSM8K, one panel per knob of \cref{tab:ablations}, as a 100-step moving average. The arms separate more here than in their held-out accuracy: the near-random orders and the weaker or stronger tilts roll out worst, yet of the four knobs $\beta$ moves the final accuracy the most.}
\label{fig:ablations}
\vspace{6pt}
\includegraphics[width=\linewidth]{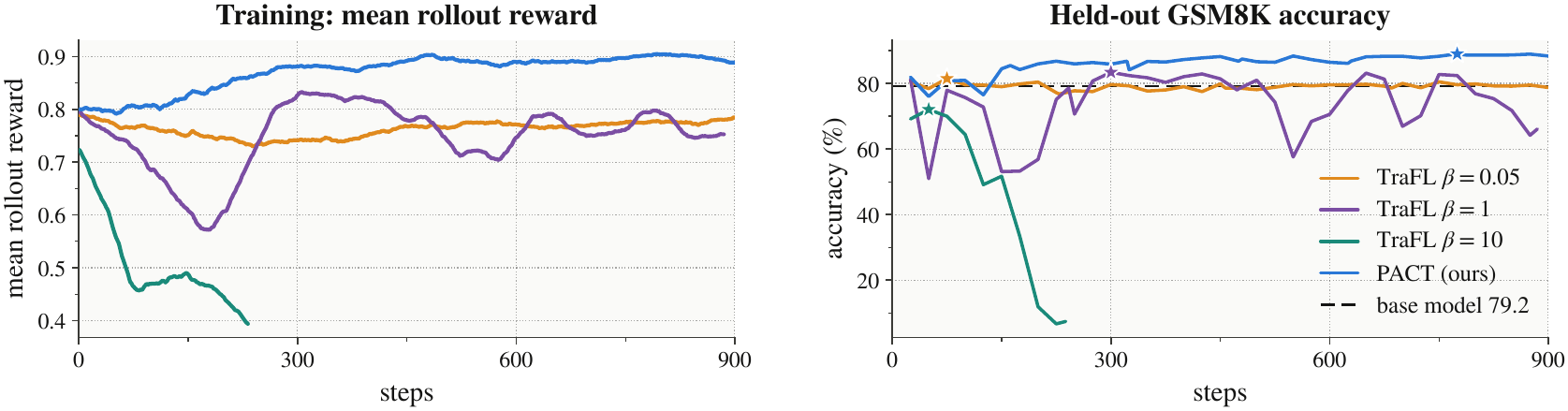}
\caption{\small \textbf{PACT outperforms TraFL, a trajectory-balance method.} TraFL at three reward sharpnesses $\beta$ against PACT, under the same rollout configuration and reward. Left: mean rollout reward, 100-step moving average. Right: held-out GSM8K accuracy at every checkpoint, with a star at each run's best. Every TraFL setting peaks within three hundred steps and then declines; PACT is still improving at step 775.}
\label{fig:trafl}
\end{figure}

\subsection{Ablations}\label{sec:ablations}

We vary one setting of the objective at a time, each arm changing a single knob from the default and seeing the same prompts at the same steps. \Cref{tab:ablations} reports held-out accuracy per knob and \cref{fig:ablations} the training reward.

PACT's accuracy is not sensitive to the number of transitions $K$ sampled per path. Across $8$, $16$, $32$ and $64$ the best held-out accuracy moves by about one binomial standard error, and only the per-step cost tracks $K$. Transitions on one path share a single boundary value and overlap heavily, so the noise to average down sits between paths, set by the number of rollouts rather than by draws within a path, and a few transitions per path suffice. We take $K{=}16$: $K{=}8$ is within one standard error of it at a fifth less time per step, and $K{=}64$ gains a third of a point, also within one standard error, at 2.2 times the step time.

The unmasking order enters through the Plackett--Luce likelihood $P^{\mathrm{ord}}_\theta(\omega_t \mid x_t)$ the sampler draws it from at temperature $\tau_o$ (\cref{sec:method}). It can also be trained explicitly, by adding the log-ratio $\gamma_{\mathrm{ord}}\log\big(P^{\mathrm{ord}}_\theta/P^{\mathrm{ord}}_{\mathrm{ref}}\big)(\omega_t \mid x_t)$ to the residual under a gate $\gamma_{\mathrm{ord}} \in \{0,1\}$. We found empirically that neither knob raises accuracy: sweeping $\tau_o$ leaves it flat from near-greedy to near-random, the near-random arm even the best despite rolling out worst, and turning the explicit term on ($\gamma_{\mathrm{ord}}{=}1$) lowers accuracy in both implementations, through the sampler's own confidences (\emph{shared}) and through a separate zero-initialized order head on detached confidences (\emph{head}). The target does not rule an order gradient out: given a finished text its law over orders is the reference's, but its per-step order kernel favours slots whose reveal raises the potential, so there is something to learn in principle. Carrying the gradient through the shared confidences lets the model change the confidence values that greedy decoding uses to schedule slots, rather than its answers; the separate head cannot affect those values and still lowers accuracy. We keep the order stochastic but ungraded ($\gamma_{\mathrm{ord}}{=}0$) at $\tau_o{=}0.2$, which rolls out as well as greedy decoding; the token-only residual then targets the token factor of the tilted kernel under the policy's own order law.

Of the four knobs, we found that the tilt $\beta$ moves held-out accuracy the most, because it has two jobs: the reward enters as $r/\beta$, so $\beta$ sets where the target sits and scales the gradient at once, which a fixed learning rate cannot separate. Weakening the tilt to $0.2$ or $0.5$ shrinks the gradient in step and under-trains within the budget ($87.3$ and $86.0$ against $89.1$), while sharpening it to $0.05$ overshoots, rollout reward falling by a quarter before recovering near step $250$, and only ties the default. We keep $\beta{=}0.1$ and do not anneal it, since $\beta$ also scales the gradient through the $r/\beta$ target, so a $\beta$ schedule at a fixed budget mostly changes the effective learning rate.

\Needspace*{12\baselineskip}
\begin{wraptable}[10]{r}{0.52\linewidth}
\vspace{-12pt}
\centering
\caption{\small \textbf{Ablations.} Best held-out GSM8K accuracy (\%) at length 256 per knob value; default in \textbf{bold}. Seconds per step for $K{=}8/16/32/64$: $32.8/41.1/58.3/92.0$.}
\label{tab:ablations}
\small
\setlength{\tabcolsep}{5pt}
\begin{tabular}{@{}lr@{\quad}lr@{\quad}lr@{\quad}lr@{}}
\toprule
$K$ & \% & $\tau_o$ & \% & $\gamma_{\mathrm{ord}}$ & \% & $\beta$ & \% \\
\cmidrule(lr){1-2}\cmidrule(lr){3-4}\cmidrule(lr){5-6}\cmidrule(lr){7-8}
$8$ & $88.4$ & $0.05$ & $88.8$ & \textbf{off} & \textbf{89.1} & $0.05$ & $89.1$ \\
$\mathbf{16}$ & $\mathbf{89.1}$ & $\mathbf{0.2}$ & $\mathbf{89.1}$ & shared & $87.7$ & $\mathbf{0.1}$ & $\mathbf{89.1}$ \\
$32$ & $88.6$ & $1$ & $88.3$ & head & $88.6$ & $0.2$ & $87.3$ \\
$64$ & $89.4$ & $5$ & $89.3$ & & & $0.5$ & $86.0$ \\
\bottomrule
\end{tabular}
\vspace{-8pt}
\end{wraptable}

  We also compare PACT with TraFL \citep{ahmadi2026beyond}, a trajectory-balance method that shares PACT's tilted target, frozen reference and squared residual but writes one residual per completion: the sequence log-ratio to the reference, a learned per-prompt normalizer and the scaled reward balance for the completion as a whole. A diffusion model has no exact sequence likelihood, so, that log-ratio is a masked-reconstruction surrogate, and the 128 transitions of a completion are constrained only through this one scalar, which lets errors on different steps cancel. We run TraFL under our rollout configuration with its objective and schedule as published and three values of $\beta$, which the paper does not specify. PACT is ahead at all three. At $\beta{=}0.05$ the residual never moves, since the per-token target gap is below the noise of the four-mask estimator, and accuracy stays at the untuned level; at $\beta{=}1$, the best setting, accuracy peaks at $83.4$ by step 300 and then declines; at $\beta{=}10$ the update drives the policy's log-ratio to the reference negative on its own samples and the reward collapses. Over the last third of training PACT leads the two usable settings by $8.6$ and $12.8$ points against a $1.4$-point standard error, and by $5.7$ at its best, while every PACT $\beta$ in \cref{tab:ablations} lands between $86.0$ and $89.1$ (\cref{fig:trafl}, \cref{app:trafl}). The comparison uses our implementation, sampler, reward and greedy Pass@1 metric, not the authors' Pass@5.

%% file: sections/06_conclusion.tex
\section{Conclusion}\label{sec:conclusion}

Most post-training methods for diffusion language models either weight a completion, or every denoising step in it, by a single completion-level advantage, or fit a reward-tilted target on completed responses or on constructed intermediate states. In this work, we propose PACT, an efficient RL method that trains the policy such that, on every transition the sampler takes, its probability ratio to the base model equals the potential gained by that step. A learned potential head carries the reward from the finished answer back to earlier states. With the ideal potential and all unmasking decisions scored, this condition is the Doob $h$-transform of the reference sampler, which reproduces the reward-tilted target; PACT's learned potential and token-only residual approximate it. The potential of the fully masked state plays the role of the normalizer, so PACT fits no separate normalizer and needs neither a sequence-likelihood estimate nor a clipped importance ratio. PACT improves on all four reasoning tasks and holds the best average accuracy across them. Further, our method also reaches AGRPO's best accuracy in about half the training compute on GSM8K. Unlike a group-relative advantage, its training signal is not identically zero when all rollouts of a prompt get the same reward, though what remains there is a consistency correction rather than reward information. Our method needs only the per-step probabilities of the recorded steps under the policy and a frozen reference, so it applies to other generators that build their output step by step.

%% file: sections/07_appendix.tex
\section{Training configurations}\label{app:configs}
\cref{tab:shared} lists PACT's configuration, one recipe for all four tasks except the learning rate; the objective's own settings are tilt $\beta{=}0.1$, order temperature $\tau_o{=}0.2$, $K{=}16$ transitions per trajectory, anchor weight $1$, $\gamma_{\mathrm{ord}}{=}0$ (token-only; the order-head variant is in \cref{sec:ablations}), and a zero-initialized two-layer potential head of hidden width 256 with learning rate $10^{-3}$ and gradient clip $5.0$, the reference being the same network with adapters disabled. AGRPO runs its released configuration \citep{zhan2025simple}, which differs in the named rows: generation length 384 in blocks of 12, random unmasking order, rank-64 LoRA, learning rate $5\times10^{-6}$, group size $N{=}8$ with group-relative advantages, clip $\epsilon{=}0.2$, and no KL term. Our AGRPO rows use the authors' released code at the commit of the current paper, with SPG's Countdown split, 384-token training, low-discrepancy timestep sampling, entropy importance sampling and the exclusion of trailing end-of-sequence steps all enabled, for 900 rollout batches, the upper end of the 400--900 steps the paper trains per task. An earlier version of the paper (\url{https://openreview.net/pdf?id=0vs1YLmmQa}) reports $40.0$ on Countdown at length 256 and the current one $79.7$; our rerun reaches $42.2$, and we have not found the source of the gap. We report the rerun because it is the only AGRPO row trained on our budget and decoded under our protocol.

\begin{table}[h]  
    \centering
\caption{\textbf{PACT training configuration}, shared across all four tasks except the learning rate.}
\label{tab:shared}
\small
\begin{tabularx}{0.82\linewidth}{@{}l >{\raggedright\arraybackslash}X@{}}
\toprule
Setting & Value \\
\midrule
base model & LLaDA-8B-Instruct \\
LoRA rank / $\alpha$ / dropout & 128 / 128 / 0.05 \\
LoRA targets & all attention + MLP projections \\
backbone precision & bfloat16 (unquantized) \\
adapters trained & LoRA only, no SFT stage \\
\midrule
optimizer & AdamW ($\beta_1{=}0.9$, $\beta_2{=}0.99$) \\
learning rate & $2\times10^{-5}$ on GSM8K, $5\times10^{-6}$ elsewhere; cosine, warmup 0.01 \\
weight decay / grad clip & 0.01 / 1.0 \\
\midrule
generation length / block & 256 / 32 \\
denoising steps per rollout & 128 \\
rollout sampling & stochastic-confidence order, $\tau_o{=}0.2$; $T{=}0.6$ ($0.9$ on Countdown), top-$p$ 0.99, top-$k$ 150 \\
rollout batches & 900 (checkpoint every 25) \\
rollouts per batch & 128 (16 prompts $\times$ 8, one per GPU) \\
devices & 16 H100, per-device batch 8 \\
\bottomrule
\end{tabularx}
\end{table}

MATH500 is the 500-problem MATH-500 subset. Evaluation uses greedy decoding ($T{=}0$) in low-confidence order with block length 32 and denoising steps equal to half the generation length, at generation lengths 256 and 512 on every benchmark, batch 8 per device; PACT's training rollouts use 256 tokens on every task. 

\section{A matched comparison with trajectory balance}\label{app:trafl}

\textbf{What TraFL optimizes.} TraFL \citep{ahmadi2026beyond} is PACT's closest published relative: the same tilted target, a frozen reference and a squared residual, but one residual per completion instead of one per transition. For a completion $y$ it minimizes $\delta(y)^2$ with
\begin{equation*}
\delta(y) \;=\; \log \hat p_\theta(y \mid q) - \log \hat p_{\mathrm{ref}}(y \mid q) - \beta\, \tilde r(y) + \log Z_\phi(q),
\end{equation*}
where $\tilde r$ is the reward centered within the rollout group and $Z_\phi$ a learned scalar per prompt. A diffusion model has no exact sequence likelihood, so $\log\hat p$ is a masked-reconstruction surrogate: the average log-probability of randomly masked tokens, i.e.\ the evidence lower bound divided by length. Every transition of a completion, 128 denoising steps in our configuration, is constrained only through this one scalar.

\textbf{A fair comparison.} Every rollout setting is that of \cref{tab:shared}: length 256 in blocks of 32, 128 denoising steps, eight rollouts for each of sixteen prompts, and the same sampler, reward and 900 batches. From TraFL we kept the objective as published: the $1/l$ normalization, four antithetic masks, non-EOS masking, a two-layer partition head on detached prompt features, one optimizer step per batch, and their $6\times10^{-5}{\to}6\times10^{-6}$ learning-rate schedule. Three differences are forced: we report greedy Pass@1 where they report Pass@5 from sixteen samples, a matched budget needs eight rollouts per prompt rather than five, and our reward is composite rather than binary. Their paper also states no numeric $\beta$, and ours cannot be reused, since their $\beta$ multiplies a per-token quantity while ours acts on a whole path. We therefore swept $\beta \in \{0.05, 1, 10\}$: $0.05$ matches our tilt per token, while $1$ and $10$ ask for about $0.7$ and $7.5$ nats per token between a correct and an incorrect completion.

\begin{table}[h]
\centering
\caption{\textbf{Trajectory balance under PACT's configuration.} Best held-out GSM8K accuracy at length 256, one decode protocol for every row. The untuned model scores $79.2$.}
\label{tab:trafl}
\small
\begin{tabular}{lc}
\toprule
Method & GSM8K (\%) \\
\midrule
TraFL, $\beta = 0.05$ & 81.5 \\
TraFL, $\beta = 1$ & 83.4 \\
TraFL, $\beta = 10$ & 72.0 \\
\midrule
PACT (ours) & \textbf{89.1} \\
\bottomrule
\end{tabular}
\end{table}

At their best checkpoints, $\beta{=}0.05$ and $\beta{=}1$ improve on the untuned model by $2.3$ and $4.2$ points. Neither keeps the gain to the end of training (\cref{fig:trafl}). $\beta{=}0.05$ peaks at step 75 and then sits at the untuned accuracy, because its per-token target gap is smaller than the noise of the four-mask estimator, so the residual never moves. $\beta{=}1$ peaks at step 300 and decays, and $\beta{=}10$ degrades from the start. Paired over the shared checkpoints past step 600, PACT leads by $8.6$ points at $\beta{=}0.05$ and $12.8$ at $\beta{=}1$, against a $1.4$-point standard error, and is ahead on every checkpoint.

One residual per completion leaves the inside of the trajectory unconstrained. The completion-level quantity moves only when all transitions move together, by an amount set by $\beta$, and the usable range of $\beta$ is narrow: too small and the signal is lost in the estimator's noise, too large and the update drives the policy's log-ratio to the reference negative on its own samples, and the reward collapses. The per-transition residual keeps $\beta$ but constrains each step, and every $\beta$ in \cref{tab:ablations} lands between $86.0$ and $89.1$. TraFL also differs from PACT in scoring completions through a masked-reconstruction surrogate rather than the on-path conditionals, and in reading its normalizer from detached prompt features, so this comparison measures the published objective as a whole. Three caveats bound this comparison: it is our implementation under our sampler, reward and metric, though its loss matches the published equations to seven decimals per completion; $\beta$ was swept at three points two orders of magnitude apart, not tuned around its best; and the authors' Pass@5 protocol was not reproduced. Under our configuration, then, trajectory balance written once per completion peaks within 300 steps and declines, while the per-transition residual is still improving at step 775.

%% file: main.bbl
\begin{thebibliography}{42}
\providecommand{\natexlab}[1]{#1}
\providecommand{\url}[1]{\texttt{#1}}
\expandafter\ifx\csname urlstyle\endcsname\relax
  \providecommand{\doi}[1]{doi: #1}\else
  \providecommand{\doi}{doi: \begingroup \urlstyle{rm}\Url}\fi

\bibitem[Ahmadi et~al.(2026)Ahmadi, Parthasarathi, and Cui]{ahmadi2026beyond}
Saba Ahmadi, Prasanna Parthasarathi, and Yufei Cui.
\newblock Beyond mode-seeking rl: Trajectory-balance post-training for
  diffusion language models.
\newblock \emph{arXiv preprint arXiv:2605.13935}, 2026.

\bibitem[Bengio et~al.(2021)Bengio, Jain, Korablyov, Precup, and
  Bengio]{bengio2021flow}
Emmanuel Bengio, Moksh Jain, Maksym Korablyov, Doina Precup, and Yoshua Bengio.
\newblock Flow network based generative models for non-iterative diverse
  candidate generation, 2021.
\newblock NeurIPS 2021.

\bibitem[Bu et~al.(2026)Bu, Huang, Han, Wong, Zhang, Suzuki, and
  Nitanda]{bu2026dprm}
Dake Bu, Wei Huang, Andi Han, Hau-San Wong, Qingfu Zhang, Taiji Suzuki, and
  Atsushi Nitanda.
\newblock {DPRM}: A plug-in token-ordering module for diffusion language
  models, 2026.
\newblock FoGen Workshop, ICML 2026.

\bibitem[Chen et~al.(2026)Chen, Wang, Potaptchik, Kim, and
  Albergo]{chen2026tilt}
Yuyuan Chen, Shiyi Wang, Peter Potaptchik, Jaeyeon Kim, and Michael~S. Albergo.
\newblock Discrete tilt matching.
\newblock \emph{arXiv preprint arXiv:2604.18739}, 2026.

\bibitem[Cobbe et~al.(2021)Cobbe, Kosaraju, Bavarian, Chen, Jun, Kaiser,
  Plappert, Tworek, Hilton, Nakano, Hesse, and Schulman]{cobbe2021training}
Karl Cobbe, Vineet Kosaraju, Mohammad Bavarian, Mark Chen, Heewoo Jun, Lukasz
  Kaiser, Matthias Plappert, Jerry Tworek, Jacob Hilton, Reiichiro Nakano,
  Christopher Hesse, and John Schulman.
\newblock Training verifiers to solve math word problems.
\newblock \emph{arXiv preprint arXiv:2110.14168}, 2021.

\bibitem[{DeepSeek-AI} et~al.(2025){DeepSeek-AI}, Guo,
  et~al.]{guo2025deepseekr1}
{DeepSeek-AI}, Daya Guo, et~al.
\newblock {DeepSeek-R1}: Incentivizing reasoning capability in {LLMs} via
  reinforcement learning.
\newblock \emph{Nature}, 645\penalty0 (8081):\penalty0 633--638, 2025.

\bibitem[Deleu et~al.(2025)Deleu, Nouri, Bengio, and Precup]{deleu2025relative}
Tristan Deleu, Padideh Nouri, Yoshua Bengio, and Doina Precup.
\newblock Relative trajectory balance is equivalent to trust-pcl, 2025.

\bibitem[Hendrycks et~al.(2021)Hendrycks, Burns, Kadavath, Arora, Basart, Tang,
  Song, and Steinhardt]{hendrycks2021measuring}
Dan Hendrycks, Collin Burns, Saurav Kadavath, Akul Arora, Steven Basart, Eric
  Tang, Dawn Song, and Jacob Steinhardt.
\newblock Measuring mathematical problem solving with the math dataset.
\newblock \emph{arXiv preprint arXiv:2103.03874}, 2021.

\bibitem[Huang et~al.(2025)Huang, Chen, Wang, Li, and Qi]{huang2025reinforcing}
Zemin Huang, Zhiyang Chen, Zijun Wang, Tiancheng Li, and Guo-Jun Qi.
\newblock Reinforcing the diffusion chain of lateral thought with diffusion
  language models, 2025.
\newblock NeurIPS 2025.

\bibitem[Labs et~al.(2025)Labs, Khanna, Kharbanda, Li, Varma, Wang, Birnbaum,
  Luo, Miraoui, Palrecha, Ermon, Grover, and Kuleshov]{labs2025mercury}
Inception Labs, Samar Khanna, Siddhant Kharbanda, Shufan Li, Harshit Varma,
  Eric Wang, Sawyer Birnbaum, Ziyang Luo, Yanis Miraoui, Akash Palrecha,
  Stefano Ermon, Aditya Grover, and Volodymyr Kuleshov.
\newblock Mercury: Ultra-fast language models based on diffusion.
\newblock \emph{arXiv preprint arXiv:2506.17298}, 2025.

\bibitem[Levine(2018)]{levine2018reinforcement}
Sergey Levine.
\newblock Reinforcement learning and control as probabilistic inference:
  Tutorial and review.
\newblock \emph{arXiv preprint arXiv:1805.00909}, 2018.

\bibitem[Lightman et~al.(2023)Lightman, Kosaraju, Burda, Edwards, Baker, Lee,
  Leike, Schulman, Sutskever, and Cobbe]{lightman2023let}
Hunter Lightman, Vineet Kosaraju, Yura Burda, Harri Edwards, Bowen Baker, Teddy
  Lee, Jan Leike, John Schulman, Ilya Sutskever, and Karl Cobbe.
\newblock Let's verify step by step.
\newblock \emph{arXiv preprint arXiv:2305.20050}, 2023.

\bibitem[Liu et~al.(2026)Liu, Xu, Stokes, Smolensky, Burger, and
  Gao]{liu2026gflowrl}
Xiaodong Liu, Michael Xu, Jack~W. Stokes, Paul Smolensky, Doug Burger, and
  Jianfeng Gao.
\newblock {GFlowRL}: Scaling distribution-matching {RL} to large language
  models.
\newblock \emph{arXiv preprint arXiv:2607.13394}, 2026.

\bibitem[Liu et~al.(2025)Liu, Xiao, Liu, Bengio, and Zhang]{liu2024diversity}
Zhen Liu, Tim~Z. Xiao, Weiyang Liu, Yoshua Bengio, and Dinghuai Zhang.
\newblock Efficient diversity-preserving diffusion alignment via
  gradient-informed {GFlowNets}.
\newblock In \emph{International Conference on Learning Representations}, 2025.

\bibitem[Malkin et~al.(2022)Malkin, Jain, Bengio, Sun, and
  Bengio]{malkin2022trajectory}
Nikolay Malkin, Moksh Jain, Emmanuel Bengio, Chen Sun, and Yoshua Bengio.
\newblock Trajectory balance: Improved credit assignment in gflownets, 2022.
\newblock NeurIPS 2022.

\bibitem[Nachum et~al.(2017)Nachum, Norouzi, Xu, and
  Schuurmans]{nachum2017bridging}
Ofir Nachum, Mohammad Norouzi, Kelvin Xu, and Dale Schuurmans.
\newblock Bridging the gap between value and policy based reinforcement
  learning.
\newblock In \emph{Advances in Neural Information Processing Systems}, 2017.

\bibitem[Nachum et~al.(2018)Nachum, Norouzi, Xu, and
  Schuurmans]{nachum2018trust}
Ofir Nachum, Mohammad Norouzi, Kelvin Xu, and Dale Schuurmans.
\newblock {Trust-PCL}: An off-policy trust region method for continuous
  control.
\newblock In \emph{International Conference on Learning Representations}, 2018.
\newblock arXiv:1707.01891.

\bibitem[Nie et~al.(2025)Nie, Zhu, You, Zhang, Ou, Hu, Zhou, Lin, Wen, and
  Li]{nie2025large}
Shen Nie, Fengqi Zhu, Zebin You, Xiaolu Zhang, Jingyang Ou, Jun Hu, Jun Zhou,
  Yankai Lin, Ji-Rong Wen, and Chongxuan Li.
\newblock Large language diffusion models, 2025.
\newblock ICML 2025.

\bibitem[Nisonoff et~al.(2024)Nisonoff, Xiong, Allenspach, and
  Listgarten]{nisonoff2024unlocking}
Hunter Nisonoff, Junhao Xiong, Stephan Allenspach, and Jennifer Listgarten.
\newblock Unlocking guidance for discrete state-space diffusion and flow
  models, 2024.
\newblock ICLR 2025.

\bibitem[Pan et~al.(2025)Pan, Zhang, Wang, Yuan, Peng, and
  Suhr]{pan2025tinyzero}
Jiayi Pan, Junjie Zhang, Xingyao Wang, Lifan Yuan, Hao Peng, and Alane Suhr.
\newblock {TinyZero}.
\newblock \url{https://github.com/Jiayi-Pan/TinyZero}, 2025.
\newblock GitHub repository.

\bibitem[Raajesh et~al.(2026)Raajesh, Shah, Klivans, and
  Kr{\"a}henb{\"u}hl]{raajesh2026mask}
Haran Raajesh, Kulin Shah, Adam Klivans, and Philipp Kr{\"a}henb{\"u}hl.
\newblock Mask-aware policy gradients for diffusion language models, 2026.
\newblock COLM 2026.

\bibitem[Rector-Brooks et~al.(2024)Rector-Brooks, Hasan, Peng, Quinn, Liu,
  Mittal, Dziri, Bronstein, Bengio, Chatterjee, Tong, and
  Bose]{rectorbrooks2024steering}
Jarrid Rector-Brooks, Mohsin Hasan, Zhangzhi Peng, Zachary Quinn, Chenghao Liu,
  Sarthak Mittal, Nouha Dziri, Michael Bronstein, Yoshua Bengio, Pranam
  Chatterjee, Alexander Tong, and Avishek~Joey Bose.
\newblock Steering masked discrete diffusion models via discrete denoising
  posterior prediction.
\newblock \emph{arXiv preprint arXiv:2410.08134}, 2024.

\bibitem[Sahoo et~al.(2024)Sahoo, Arriola, Schiff, Gokaslan, Marroquin, Chiu,
  Rush, and Kuleshov]{sahoo2024simple}
Subham~Sekhar Sahoo, Marianne Arriola, Yair Schiff, Aaron Gokaslan, Edgar
  Marroquin, Justin~T Chiu, Alexander Rush, and Volodymyr Kuleshov.
\newblock Simple and effective masked diffusion language models, 2024.
\newblock NeurIPS 2024.

\bibitem[Shao et~al.(2024)Shao, Wang, Zhu, Xu, Song, Bi, Zhang, Zhang, Li, Wu,
  and Guo]{shao2024deepseekmath}
Zhihong Shao, Peiyi Wang, Qihao Zhu, Runxin Xu, Junxiao Song, Xiao Bi, Haowei
  Zhang, Mingchuan Zhang, Y.~K. Li, Y.~Wu, and Daya Guo.
\newblock Deepseekmath: Pushing the limits of mathematical reasoning in open
  language models.
\newblock \emph{arXiv preprint arXiv:2402.03300}, 2024.

\bibitem[Singh et~al.(2026{\natexlab{a}})Singh, No{\"e}l, Scholak, Belilovsky,
  and Ostapenko]{singh2026fixed}
Vaibhav Singh, Pierre-Andr{\'e} No{\"e}l, Torsten Scholak, Eugene Belilovsky,
  and Oleksiy Ostapenko.
\newblock Fixed state, long reach: What a constant-size cache buys block
  diffusion at scale.
\newblock \emph{arXiv preprint arXiv:2609.11998}, 2026{\natexlab{a}}.

\bibitem[Singh et~al.(2026{\natexlab{b}})Singh, Ostapenko, Noel, Belilovsky,
  and Scholak]{singh2026diffumamba}
Vaibhav Singh, Oleksiy Ostapenko, Pierre-Andre Noel, Eugene Belilovsky, and
  Torsten Scholak.
\newblock Diffumamba: High-throughput diffusion {LM}s with mamba backbone.
\newblock In \emph{Forty-third International Conference on Machine Learning},
  2026{\natexlab{b}}.
\newblock URL \url{https://openreview.net/forum?id=hLizFuTBsA}.

\bibitem[So et~al.(2026)So, Karrer, Fan, Chen, and Liu]{so2026discrete}
Oswin So, Brian Karrer, Chuchu Fan, Ricky T.~Q. Chen, and Guan-Horng Liu.
\newblock Discrete adjoint matching, 2026.
\newblock ICLR 2026.

\bibitem[Song et~al.(2026)Song, You, Tian, Su, Huang, and Hou]{songdiffureason}
Yiping Song, Jinyu You, Zhiliang Tian, Jinsong Su, Minlie Huang, and Chenping
  Hou.
\newblock Diffureason: Enhancing reasoning ability for diffusion language
  models via monte carlo tree search.
\newblock In \emph{Forty-third International Conference on Machine Learning},
  2026.

\bibitem[Svete \& Sabharwal(2026)Svete and Sabharwal]{svete2026reasoning}
Anej Svete and Ashish Sabharwal.
\newblock On the reasoning abilities of masked diffusion language models.
\newblock In \emph{International Conference on Learning Representations},
  volume 2026, pp.\  139646--139681, 2026.

\bibitem[Tang et~al.(2025)Tang, Dolga, Yoon, and Bogunovic]{tang2025wd1}
Xiaohang Tang, Rares Dolga, Sangwoong Yoon, and Ilija Bogunovic.
\newblock wd1: Weighted policy optimization for reasoning in diffusion language
  models, 2025.
\newblock ICLR 2026.

\bibitem[Tang et~al.(2026)Tang, Jiang, Liu, Zhao, Xu, Yoon, and
  Bogunovic]{tang2026gdsd}
Xiaohang Tang, Keyue Jiang, Che Liu, Qifang Zhao, Xiaoxiao Xu, Sangwoong Yoon,
  and Ilija Bogunovic.
\newblock Gdsd: Reinforcement learning as guided denoiser self-distillation for
  diffusion language models.
\newblock \emph{arXiv preprint arXiv:2605.29398}, 2026.

\bibitem[Venkatraman et~al.(2024)Venkatraman, Jain, Scimeca, Kim, Sendera,
  Hasan, Rowe, Mittal, Lemos, Bengio, Adam, Rector-Brooks, Bengio, Berseth, and
  Malkin]{venkatraman2024amortizing}
Siddarth Venkatraman, Moksh Jain, Luca Scimeca, Minsu Kim, Marcin Sendera,
  Mohsin Hasan, Luke Rowe, Sarthak Mittal, Pablo Lemos, Emmanuel Bengio,
  Alexandre Adam, Jarrid Rector-Brooks, Yoshua Bengio, Glen Berseth, and
  Nikolay Malkin.
\newblock Amortizing intractable inference in diffusion models for vision,
  language, and control, 2024.

\bibitem[Wang et~al.(2025{\natexlab{a}})Wang, Rashidinejad, Su, Jiang, Wang,
  Zhao, Zhou, Shen, Chen, Jaakkola, Tian, and Liu]{wang2025spg}
Chenyu Wang, Paria Rashidinejad, DiJia Su, Song Jiang, Sid Wang, Siyan Zhao,
  Cai Zhou, Shannon~Zejiang Shen, Feiyu Chen, Tommi Jaakkola, Yuandong Tian,
  and Bo~Liu.
\newblock Spg: Sandwiched policy gradient for masked diffusion language models,
  2025{\natexlab{a}}.
\newblock ICLR 2026.

\bibitem[Wang et~al.(2025{\natexlab{b}})Wang, Rashidinejad, Su, Jiang, Wang,
  Zhao, Zhou, Shen, Chen, Jaakkola, Tian, and Liu]{wang2025spgrepo}
Chenyu Wang, Paria Rashidinejad, DiJia Su, Song Jiang, Sid Wang, Siyan Zhao,
  Cai Zhou, Shannon~Zejiang Shen, Feiyu Chen, Tommi Jaakkola, Yuandong Tian,
  and Bo~Liu.
\newblock {SPG}: {S}andwiched {P}olicy {G}radient for {M}asked {D}iffusion
  {L}anguage {M}odels.
\newblock \url{https://github.com/facebookresearch/SPG}, 2025{\natexlab{b}}.
\newblock GitHub repository; we use its $4\times4$ Sudoku data and prompt
  setup.

\bibitem[Yang et~al.(2025)Yang, Tian, Li, Zhang, Shen, Tong, and
  Wang]{yang2025mmada}
Ling Yang, Ye~Tian, Bowen Li, Xinchen Zhang, Ke~Shen, Yunhai Tong, and Mengdi
  Wang.
\newblock Mmada: Multimodal large diffusion language models, 2025.
\newblock NeurIPS 2025.

\bibitem[Ye et~al.(2024)Ye, Gong, Chen, Zheng, Gao, Shi, Wu, Jiang, Li, Bi, and
  Kong]{ye2024diffusion}
Jiacheng Ye, Shansan Gong, Liheng Chen, Lin Zheng, Jiahui Gao, Han Shi, Chuan
  Wu, Xin Jiang, Zhenguo Li, Wei Bi, and Lingpeng Kong.
\newblock Diffusion of thoughts: Chain-of-thought reasoning in diffusion
  language models, 2024.
\newblock NeurIPS 2024.

\bibitem[Ye et~al.(2025)Ye, Xie, Zheng, Gao, Wu, Jiang, Li, and
  Kong]{ye2025dream}
Jiacheng Ye, Zhihui Xie, Lin Zheng, Jiahui Gao, Zirui Wu, Xin Jiang, Zhenguo
  Li, and Lingpeng Kong.
\newblock Dream 7b: Diffusion large language models.
\newblock \emph{arXiv preprint arXiv:2508.15487}, 2025.

\bibitem[Yu et~al.(2026)Yu, Xu, Jiang, Zhang, and Zou]{yu2026relative}
Zichao Yu, Shengze Xu, Bingqing Jiang, Wenyi Zhang, and Difan Zou.
\newblock Relative score policy optimization for diffusion language models.
\newblock \emph{arXiv preprint arXiv:2605.10218}, 2026.

\bibitem[Zhan(2025)]{zhan2025simple}
Anthony Zhan.
\newblock Simple policy gradients for reasoning with diffusion language models.
\newblock \emph{arXiv preprint arXiv:2510.04019}, 2025.

\bibitem[Zhao et~al.(2025)Zhao, Gupta, Zheng, and Grover]{zhao2025d1}
Siyan Zhao, Devaansh Gupta, Qinqing Zheng, and Aditya Grover.
\newblock d1: Scaling reasoning in diffusion large language models via
  reinforcement learning, 2025.
\newblock NeurIPS 2025.

\bibitem[Zhu et~al.(2025{\natexlab{a}})Zhu, Cheng, Zhang, Li, Zhang, Jiang,
  Sun, Hua, Zuo, Lv, Zhang, Chen, Shao, Xue, Song, Yang, Cui, Ding, Gao, Liu,
  Zhou, Mei, and Lin]{zhu2025flowrl}
Xuekai Zhu, Daixuan Cheng, Dinghuai Zhang, Hengli Li, Kaiyan Zhang, Che Jiang,
  Youbang Sun, Ermo Hua, Yuxin Zuo, Xingtai Lv, Qizheng Zhang, Lin Chen,
  Fanghao Shao, Bo~Xue, Yunchong Song, Zhenjie Yang, Ganqu Cui, Ning Ding,
  Jianfeng Gao, Xiaodong Liu, Bowen Zhou, Hongyuan Mei, and Zhouhan Lin.
\newblock Flowrl: Matching reward distributions for llm reasoning,
  2025{\natexlab{a}}.
\newblock ICLR 2026.

\bibitem[Zhu et~al.(2025{\natexlab{b}})Zhu, Guo, Choi, Molodyk, Yuan, Tao, and
  Chen]{zhu2025enhancing}
Yuchen Zhu, Wei Guo, Jaemoo Choi, Petr Molodyk, Bo~Yuan, Molei Tao, and Yongxin
  Chen.
\newblock Enhancing reasoning for diffusion llms via distribution matching
  policy optimization.
\newblock \emph{arXiv preprint arXiv:2510.08233}, 2025{\natexlab{b}}.

\end{thebibliography}
